\documentclass[letterpaper, 10 pt, conference]{ieeeconf}  

\IEEEoverridecommandlockouts                              

\usepackage{textcomp, libertine}
\usepackage{cite}
\usepackage{amsmath,amssymb,amsfonts}
\usepackage{algorithmic}
\usepackage{algorithm,mathtools}
\usepackage{graphicx}
\usepackage{textcomp}
\usepackage{epsfig} 
\usepackage{mathptmx} 
\usepackage{times} 
\usepackage[T1]{fontenc}
\usepackage[utf8]{inputenc}
\usepackage{float}
\usepackage{epstopdf}
\usepackage{textcomp}
\usepackage{multirow}
\let\labelindent\relax
\usepackage{enumitem}
\usepackage{color}
\usepackage{soul}
\usepackage[caption=false]{subfig}
\usepackage{caption,setspace}
\usepackage{hyperref}

\hypersetup{
    colorlinks=false,
    linkcolor=black,
    filecolor=magenta,      
    urlcolor=blue,
}
\title{\LARGE \bf
iART: Learning from Demonstration for Assisted Robotic Therapy Using LSTM}

\author{Shrey Pareek$^{1}$ and Thenkurussi Kesavadas$^{2}$
\thanks{Research supported by the National
Science Foundation under Grant No. 1502339.}
\thanks{$^{1}$ Shrey Pareek is with the Department of Industrial and Enterprise Systems Engineering, University of Illinois,
        Urbana-Champaign, IL-USA,
        {\tt\small spareek2@illinois.edu}}%
\thanks{$^{2}$ Thenkurussi Kesavadas is with the Department of Industrial and Enterprise Systems Engineering, University of Illinois,
        Urbana-Champaign, IL-USA,
        {\tt\small kesh@illinois.edu}}%
}

\begin{document}

\maketitle
\thispagestyle{empty}
\pagestyle{empty}

\begin{abstract}
In this paper, we present an intelligent Assistant for Robotic Therapy (iART), that provides robotic assistance during 3D trajectory tracking tasks.  We propose a novel LSTM-based robot learning from demonstration (LfD) paradigm to mimic a therapist's assistance behavior. iART presents a trajectory agnostic LfD routine that can generalize learned behavior from a single trajectory to any 3D shape. Once the therapist's behavior has been learned, iART enables the patient to modify this behavior as per their preference. The system requires only a single demonstration of 2 minutes and exhibits a mean accuracy of 91.41\% in predicting, and hence mimicking a therapist's assistance behavior. The system delivers stable assistance in realtime and successfully reproduces different types of assistance behaviors.
\end{abstract}
\section{INTRODUCTION}\label{sec:intro}

Traditional stroke therapy is hospital-centric and involves a team of doctors and therapists designing repetitive therapy exercises for a patient. The success of these programs relies heavily on the ability of the patient to practice the prescribed exercises at home. In these situations, robot-based tele-rehabilitation systems serve as reliable tools for home-based therapy. 

Robot-based tele-rehabilitation systems \cite{brewer2007poststroke} comprise of a robotic system that can assist/resist a patient's motions as they perform therapy exercises using a simulation system. These systems also allow a remote therapist to observe and assist the patient over a network and modify the therapy tasks and/or robotic assistance/resistance \cite{pareek2018development}.

The robotic device may be programmed to automatically assist/resist the patient in the absence of a therapist. The selection of the type of robotic-assistance is a crucial yet challenging aspect of robotic rehabilitation. A task rendered too difficult due to the presence of inadequate robotic assistance may induce anxiety \cite{nakamura2014concept} in the patient; forcing them to quit the rehabilitation task early. Conversely, excessive assistance in a task may not challenge the patient enough and can lead to boredom \cite{nakamura2014concept}. Further, this may lead to over-reliance on the robotic assistance which may be inhibitory to the rehabilitation outcome \cite{Wolbrecht2008, Esfahani2015}. Thus, the choice of the assistance mechanism is crucial to the success of the rehabilitative routine.


One such commonly used assistance mechanism is the error reduction (ER) strategy. ER  refers to an assistance strategy that minimizes the tracking error in a trajectory tracking task. These strategies are usually dependent on the instantaneous tracking error and follow strict threshold-based rules. In other words, this strategy constrains the subject within a fixed no-error zone around the reference trajectory. If the subject deviates outside this zone, the robotic device provides a corrective force and guides them back inside this zone. Mathematically, if the tracking error is larger than a threshold $r$ (called the radius of the no-error zone), robotic assistance is switched on. Conversely, if the tracking error is lower than this threshold, assistance is switched off (see Appendix for details).  Such rule-based methods only use the instantaneous position error as a metric for deciding whether or not to assist the patient; and do not take into account the patient's performance history. This makes the selection of the size of the no-error zone, very crucial and challenging.

\begin{figure}[!t]
	\centering
	\includegraphics[ width=.45\textwidth]{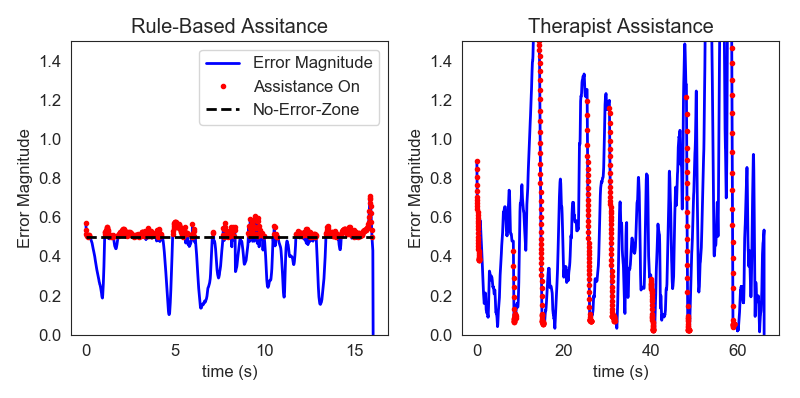}
	\caption{Tracking behavior demonstrated under rule-based assistance (left) and therapist assistance (right).}
	\vspace{-.5cm}
	\label{fig:comparison}
\end{figure}

In our experiments with robot-based trajectory tracking, we observed that subjects tend to stay at the boundary of the virtual no-error zone, as at the boundary, the haptic device provides minimal assistance enabling the subjects to correctly follow the trajectory with minimal effort and low tracking error. 
The plots in Fig. \ref{fig:comparison} show the absolute tracking error v/s time (blue line) exhibited by a subject under two assistance conditions (ER and human therapist) for the same reference trajectory. 
The red dots denote instances when the robotic assistance was switched on; and the dotted black line represents the radius ($r$) of the no-error zone (see Appendix). 

It is evident that under the rule-based assistance, the subject tends to stay close to the boundary of the no-error zone. The robotic assistance is switched on for around $50\%$ ($\sim7.5s$)  of the duration of the experiment. In the second case, the therapist assists the subject for about $10\%$ ($\sim6.5s$) of the experiment and allows them to reduce the tracking error on their own accord before intervening.  

Further, under ER, the subject completes the tracking task in 15 seconds whereas they require a little more than a minute in the second case for the same task. Conventional wisdom dictates that lower task completion time signifies superior performance. However, it is evident that the lower completion time observed in the first case is due to the presence of excessive robotic assistance. This increased dependency on the robotic assistance may be inhibitory to the therapy outcome \cite{Wolbrecht2008}.

In other words, although rule-based approaches might improve patient performance during assisted training, they may hinder their ability to learn the desired skills for successful task completion in the absence of this assistance. 
Hence, the choice of when to assist the patient needs be made carefully to ensure that skills learned during robot assisted rehabilitation are transferred to real-world scenarios. 

\section{ROBOT LEARNING FROM DEMONSTRATION}

We propose the use of a robot learning from demonstration (LfD)  strategy to learn a therapist's assistance behavior (see Fig. \ref{fig:comparison}). A key challenge in robotics is the problem of learning a mapping between the state space and actions. This mapping, commonly referred to as policy, describes a set of rules that enables a robot to select an optimal action based on its current state. Although these rules or policies can be generated by hand (rule-based methods); this process can be exhaustive and usually only works reliably for certain scenarios \cite{argall2009survey}. 

LfD refers to a methodology wherein the policy is learned from demonstrations or examples provided by a teacher (in case of robotic rehabilitation the therapist is the teacher). These examples refer to a set of state-action pairs recorded during a \textit{demonstration phase} - wherein the teacher demonstrates the desired robot behavior. LfD algorithms then use these examples to derive a policy that reliably reproduces the demonstrated behavior. LfD does not require a large library of state-action pairs as the demonstrations can be focused to areas of the state-space that will actually be encountered during the task execution. Finally, since LfD relies on the assumption that expert demonstrations are based on the optimal policy to be reproduced; the learning process is faster.

In this paper, we approach LfD as a supervised learning problem to learn a therapist's assistance behavior through a set of state-action pairs. We propose an \textbf{i}ntelligent \textbf{A}ssistant for \textbf{R}obotic \textbf{T}herapy (iART), that enables home-based robotic therapy for stroke rehabilitation. iART refers to a LSTM-based \cite {gers1999learning} LfD paradigm that monitors patient-therapist interactions and generates a patient-centric assistance platform using a single 2-minute trajectory tracking task. The system does not require any external sensors such as brain-computer-interface \cite{pareek2015cognitive} or surface electromyography \cite{pareek2019myotrack}. The proposed method is trajectory agnostic, which implies it can learn from a single trajectory and be generalized for multiple scenarios. Use of LSTM takes into account the historical performance of the patient while providing assistance, leading to the realization of an intuitive assistive paradigm. Finally, iART enables the patient to modify the behavior of the system as per their requirement using an iterative learning process. A video demo of iART may be found \href{https://youtu.be/Gc-VVE-lJaA}{\textit{here}}\footnote{https://youtu.be/Gc-VVE-lJaA}.

\subsection{LfD in Robotic Rehabilitation} \label{sec:literature}

Although LfD has been widely implemented in mainstream robotics \cite{argall2009survey}, it is a relatively nascent concept in the field of assisted stroke rehabilitation. Tavakoli's research group \cite{maaref2016bicycle, najafi2017robotic, fong2019therapist} have pioneered the use of LfD in robotic rehabilitation. Their algorithms rely on learning the \textit{average} impedance and/or position tracking behavior exhibited by an expert while conducting a therapy task. Here, \textit{average} position tracking refers to the tracking behavior obtained over multiple trials of the \textit{same} trajectory tracking task. Similarly, the average impedance behavior can be used to model the kinesthetic behavior of a therapist \cite{fong2019therapist}.

This average behavior is then used as a reference to which the patient's performance in realtime is compared. Any deviations from this reference are used to adjust the task difficulty/assistance for the patient \cite{maaref2016bicycle}. Jung et al. \cite{jung2015learning} use Latent Dirichlet Allocation to model demonstrations of a therapist in the form of a generative process for autonomous therapy. However, this methodology is task/trajectory specific and the therapist needs to train the system for multiple different trajectories in order to develop a practically viable system. This makes the demonstration process highly exhaustive and limits the scalability of the system. 


Additionally, the above methods only learn the average behavior of the therapist and do not take into account the preferences of the patient. While adherence to the therapy regime prescribed by the therapist is essential for successful rehabilitation; an ideal system should focus on learning the therapist-patient interaction rather than just the therapist's behavior.

iART makes two critical distinctions from the above methodologies. First, instead of learning the average therapist behavior and providing assistance based on deviations from this average behavior, iART learns when a therapist chooses to assist a patient as they perform a trajectory tracking task. This enables iART to be trajectory agnostic, making it scalable. However, this means that iART does not learn the impedance behavior of a therapist \cite{maaref2016bicycle, fong2019therapist} and only learns 'when-to-assist' and not 'how-to-assist'. This is a limitation of the proposed methodology. Second, iART focuses on observing therapist-patient interactions rather than just the therapist's behavior. This enables iART to be tuned as per the specific requirements of a patient. However, this requires individual demonstrations for every patient. We now describe iART's algorithm in detail.
\begin{figure*}[!t]
	\centering
	\includegraphics[width=\textwidth]{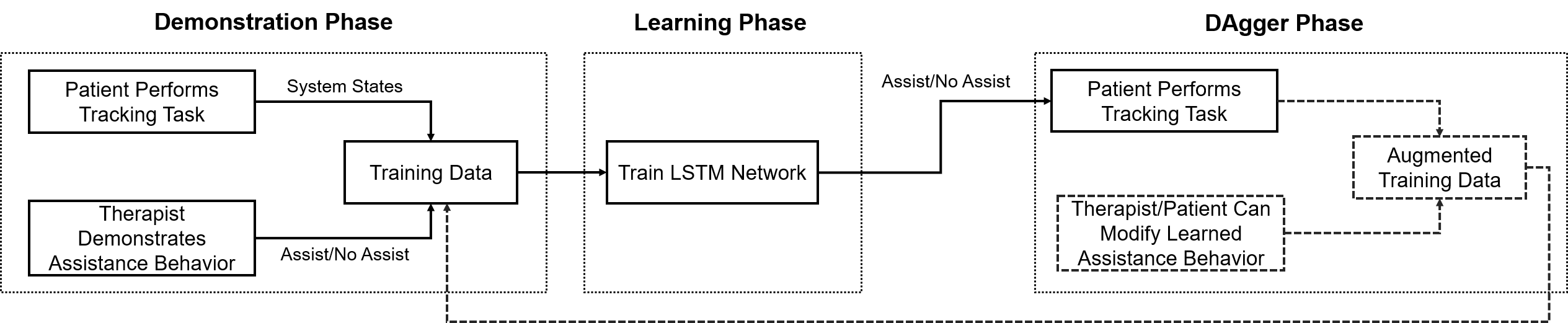}
	\caption{Schematic representation of iART.}
	\vspace{-.5cm}
	\label{fig:schematic_lstm}
\end{figure*}

\section{\lowercase{i}ART}
\begin{figure}[!t]
	\centering
	\includegraphics[width=.45\textwidth]{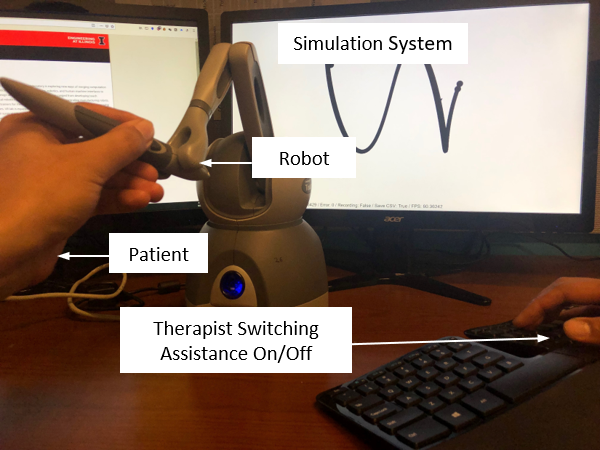}
	\caption{Experimental setup.}
	\vspace{-.5cm}
	\label{fig:setup}
\end{figure}

iART uses LSTM to learn a therapist's assistance behavior. LSTM is a class of neural networks that takes into account the past history through a sequence of features. LSTM has the ability to memorize information over a series of timesteps using the concept of memory cells. $c_k \in \mathbb{R}^n$ represents this long term memory at timestep $k$. $\hat{c}_{k}$ represents a candidate state at $k$. During training LSTM learns whether to overwrite a memory cell, retrieve it, or keep it for the next time
step. The formulation for the output state $h$ at a future timestep $k+1$ is given by \cite{amrouche2018long}:

\begin{eqnarray}\label{eq:lstm}
\nonumber
    \hat{c}_{k+1} &=& tanh(w_c[h_k,x_{k+1}],b_c)\\\nonumber
    c_{k+1} &=&  \sigma(w_f[h_{k},x_{k+1}],b_f)\odot c_k + \sigma(w_i[h_{k},x_{k+1}],b_i)\odot \hat{c}_{k+1}\\
    h_{k+1} &=&\sigma(w_o[h_{k},x_{k+1}],b_o) \odot tanh(c_{k+1})
\end{eqnarray}

where, $\tanh(\cdot)$ is the tangent hyperbolic function and $\sigma(\cdot) = \exp(\cdot)/(1 + \exp(\cdot))$ is the sigmoid function. $i, f,$ and $o$ refer to the input, forget and output gates, respectively. $W_x$ and $b_x$ are the weights and biases associated with these gates. The symbol $\odot$ denotes the Hadamard product. 

Traditionally, LSTM finds applications in scenarios such as natural language processing and time-series prediction. We propose a novel paradigm of using LSTM in LfD for robotic rehabilitation. Consider the LfD methodologies described in Section \ref{sec:literature}. These methods rely on the instantaneous behavior of the patient in order to decide the degree of assistance to be supplied. If the system observes deviations from the average behavior, it immediately supplies robotic assistance to the patient. This greedy approach does not allow the patient to reduce the tracking errors on their own accord and may increase the patient's dependence on the robot. 

On the other hand, a therapist assisting the patient observes the patient's behavior over a brief period of time before supplying any assistance. They may allow the patient to return to the desired trajectory autonomously. If however, the patient continues to struggle and cannot execute the task properly, the therapist may decide to intervene and assist the patient. Use of LSTM enables iART to capture this history-based patient-therapist interaction and successfully reproduce it for the realization of an intuitive assistive paradigm. Non-history-based classifiers such as support vector machine or neural networks provide assistance based on the instantaneous performance of the subjects. In our initial experiments, LSTM outperformed all other types of non-history-based classifiers. Overall, iART's architecture comprises of three key phases (Fig. \ref{fig:schematic_lstm}):
\begin{enumerate}
    \item Demonstration Phase: refers to the data collection phase wherein the patient performs the tracking task using a robotic device and the therapist decides whether or not to supply robotic assistance to the patient based on their performance. 
    \item Learning Phase: involves the use of an LSTM-based neural network to learn the therapist's assistance behavior observed during the demonstration phase.
    \item DAgger Phase: enables the patient/therapist to test the realtime assistance behavior learned by iART and iteratively change the behavior if needed.
\end{enumerate}

We now describe these modules individually. 

\subsection{Demonstration Phase} \label{sec:demo}

The demonstration phase involves a shared therapist-patient interaction session during which a 3D reference trajectory is chosen. The patient is then required to track this reference path using the end-effector of a robotic device to the best of their ability (Fig. \ref{fig:setup}). By default, the robot does not offer any assistance during the execution of these trajectories. If the expert deems that robotic assistance needs to be applied, they can instruct the robot (using a keyboard key press) to guide the subject back to the closest point on the trajectory (denoted by $\mathbf{x_l}$  in Fig. \ref{fig:errorZones}). 
Assistance is provided in accordance with the ER rule described via (\ref{eq:control}). iART eliminates the 'no-error zone' parameter ($r$) and instead assists the subject to the closest point on the trajectory ($x_l$). In other words, $r$ can be set to zero and the error threshold that decides whether assistance needs to be provided is replaced by the non-linear LSTM function that depends on multiple performance measures or features (described later).

Once the therapist decides that adequate assistance has been provided to the patient, they can disable the robotic assistance using another key press. On average, the demonstration phase lasts for 2 minutes. Switching the assistance on would lead to a sudden jump in the robotic assistance from 0 to a high value, and may 'jerk' the subject's arm around. The $K_p$ gain in (\ref{eq:control}) was chosen so as to minimize this effect while providing adequate assistance. In the future however, the gain would be gradually increased from zero to a maximum value (and vice versa) to completely eliminate this jump in robotic force. 
    
During this phase, various local states are logged at a sampling rate of 30 Hz. These states are - tracking error in x ($e_x$), y ($e_y$), and z ($e_z$) direction, error magnitude ($e$), radius of curvature ($r_c$), velocity magnitude ($v$), and a boolean describing whether the subject is tracking the trajectory or navigating to the start point of the trajectory after completion of a trial ($is_{track}$). Mathematically, the states are denoted by a vector $\mathbf{s} = [e_x,e_y,e_z,e,r_c,v,is_{track}]$.

\subsubsection{State-Action Pairs}

The use of local states enables us to train a trajectory agnostic LfD model. Consider the use of x, y, z error. Using tracking error as a feature as opposed to the absolute position of the robot end-effector is analogous to using distance information from LIDAR or RADAR signals to localize neighboring vehicles during autonomous driving. The assistance behavior depends on the distance from the reference trajectory and not the location of the end-effector in the world-space.  Velocity magnitude captures the ability of a patient to reduce any tracking errors. For instance, a lower velocity with high tracking error may denote the inability of the patient to return the end effector closer to the reference trajectory. 
Use of radius of curvature captures the assistance behavior w.r.t. the complexity of the trajectory. Straight sections of the trajectory (large radius of curvature) are generally easier to traverse and would exhibit lower tracking error; signifying need of assistance even when the error magnitude may be small. Conversely, curved sections (small radius of curvature) would allow more leeway in terms of the allowable error before the assistance in enabled. 

It should be noted that the effects of the states such as velocity magnitude or curvature on assistance described above have been only included as an intuition to explain the reasoning behind the choice of these features. Since we are using an LSTM (with highly non-linear dynamics), it is impossible to identify the true role that these states might play during the actual learning process. 
Local states at each epoch are associated with a corresponding action - robotic assistance \textit{On} or \textit{Off}. The action is denoted by $a\in\{0,1\}$. This set of state-action pairs ($\mathbf{s},a$) is used for supervised learning during the learning phase.

\subsection{Learning Phase} \label{sec:learning}


 Learning phase refers to the training phase of iART wherein an LSTM model is used to mimic the therapist-patient interaction recorded during the demonstration phase. The learned  model is then used to  automatically supply realtime adaptive assistance to the patient in the absence of the therapist. 

The LSTM architecture comprises of one hidden LSTM layer with 100 neurons, each with a hyperbolic tangent (tanh) activation function. The output layer uses a single neuron with sigmoid activation function. A batch size of 32 is used. Finally, a mean-squared-error (MSE) loss function is used with the Adam optimizer and the model is trained for 400 epochs.

As mentioned earlier, the network uses a set of local state-action pairs as the labelled dataset for supervised learning. Since LSTM requires historic data, we use a moving window of 1 sec (or 30 data points, since the data are sampled at a rate of 30 Hz) to provide this past information. This means that the network observes the patient behavior over a moving window of 1 sec (with 96.66\% or a 29 step overlap between subsequent windows) to determine whether or not to provide assistance to the subject. The final output is given as the probability of assistance ($P(A)$). If $P(A)>0.5$, robotic assistance is supplied to the patient and vice-versa. 

\subsection{DA\lowercase{gger} Phase}\label{sec:dagger}
The focus of iART is the development of a patient-centric assistance platform. To this end, iART enables the patient (or therapist) to modify the assistance behavior learned by iART as per their need through iART's DAgger module. 

The DAgger module is inspired by the Dataset Aggregation (DAgger) methodology proposed in \cite{ross2011reduction}. DAgger is an iterative imitation learning algorithm that aims at mimicking the expert's policy $\pi_e$, iteratively. Briefly, in the first iteration, a dataset of expert examples $\chi$ is collected and used to extract the approximate policy $\hat{\pi_1}$. Then at iteration $n$, it uses policy $\hat{\pi_{n-1}}$ (i.e. the previous best policy) to collect more examples and adds them to the dataset $\chi$. The augmented dataset is then used to estimate the next policy $\pi_{n}$ that best mimics the expert's policy $\pi_e$ \footnote{Detailed explanation of DAgger may be found in the original implementation at \cite{ross2011reduction}}. 

\subsubsection{Modifying DAgger For Retraining}

In this paper, we modify the original DAgger implementation to enable a patient  to change the assistance behavior of pre-trained models. For instance, consider the straight line tracking task between two points $\mathbf{p_1}$ and $\mathbf{p_2}$ as shown in Fig. \ref{fig:errorZones}. While the patient tracks the straight line between $\mathbf{p_1}$ and $\mathbf{p_2}$, the therapist decides whether or not to provide assistance. Once the patient navigates to the end point of the line ($\mathbf{p_2}$), they are required to return to the starting point ($\mathbf{p_1}$) and continue the tracking task. The therapist may choose to assist the patient as they traverse their path back to the starting point. In this case, if the robotic assistance is switched on, the patient will be guided directly to $\mathbf{p_1}$ \footnote{ In other words, the desired point $\mathbf{x_d}$ in (\ref{eq:control}) will be set as $\mathbf{p_1}$.}.

Consider the following scenario - during the demonstration phase (Section \ref{sec:demo}) the therapist always assists the patient as they return to the starting point $\mathbf{p_1}$. The LSTM model described in Section \ref{sec:learning} uses a dataset $\chi$ that exhibits this behavior to learn a representation of this policy $\hat{\pi_1}$. This is analogous to the first iteration of DAgger described above. Over time the patient's performance improves and they no longer require assistance while returning to $\mathbf{p_1}$. 

The DAgger module enables the patient to modify this behavior. The patient can override iART's actions by pressing a key on the keyboard. In this case, since iART will assist the patient by guiding them back to $\mathbf{p_1}$, the patient overrides this action to not assist during the return phase. These overridden data points $\chi_o$ are then used to augment the original dataset $\chi$. This is a modification of the original DAgger implementation, wherein all data points are used to augment the dataset. 

The LSTM network in then retrained on the augmented dataset. Additionally, the overridden data points $\chi_o$ are assigned higher weightage during the learning task to adapt quickly to the new desired behaviour. We use a ratio of $\beta$:1 ($\beta=20$ in this case) to assign higher weightage to the overridden data points. This means that, mis-classifications on the overridden data points $\chi_o$, are penalized  $\beta$ times more by the loss function ( $\beta^2$ times in case of MSE) compared to the data points in the original dataset $\chi$. The new policy $\pi_n$ is then learned on the augmented dataset (see Algorithm \ref{alg:iart}). 

\begin{algorithm}[!h] 
 \caption{iART Algorithm: Learning Expert Policy $\pi_e$}
\label{alg:iart}
\begin{algorithmic}[1]
    \STATE Initialize $\chi \leftarrow \phi$
    \STATE \textbf{Demonstration Phase:} 
    \begin{ALC@g}
        \STATE Collect expert state-action pairs $[\mathbf{s},a]\rightarrow \chi$.
    \end{ALC@g}
    \STATE \textbf{Learning Phase:} 
    \begin{ALC@g}
        \STATE Train LSTM on $\chi$ to learn approximate policy $\hat{\pi_1}$.
    \end{ALC@g}
    \STATE \textbf{DAgger Phase:} 
    \begin{ALC@g}
        \FOR{$n = 2,3,\dots$}
        \STATE Get dataset $\chi_o$ of overridden state-action pairs.
        \STATE Aggregate data sets: $\chi\leftarrow\chi\cup\beta\chi_o$.
        \STATE \textbf{Learning Phase:} 
        \begin{ALC@g}
            \STATE Learn new policy $\hat{\pi_n}$ using LSTM on updated $\chi$.
        \end{ALC@g}
        \ENDFOR
    \end{ALC@g}
    \STATE \textbf{Return} chosen policy $\hat{\pi_i}$ for $i\in1,2,\dots$.
\end{algorithmic}
\end{algorithm}

 \section{EXPERIMENTAL EVALUATION}
 
We studied the efficacy and feasibility of iART through four different experiments. 
 Nine healthy subjects (average age 27.22 years; 7 male; 2 female; 8 right handed; 1 left handed) who served as the 'patient' were recruited for the experiments. One of the authors served as the 'therapist'. 
 
\begin{figure}[b!]
	\centering
	\includegraphics[width=.45\textwidth]{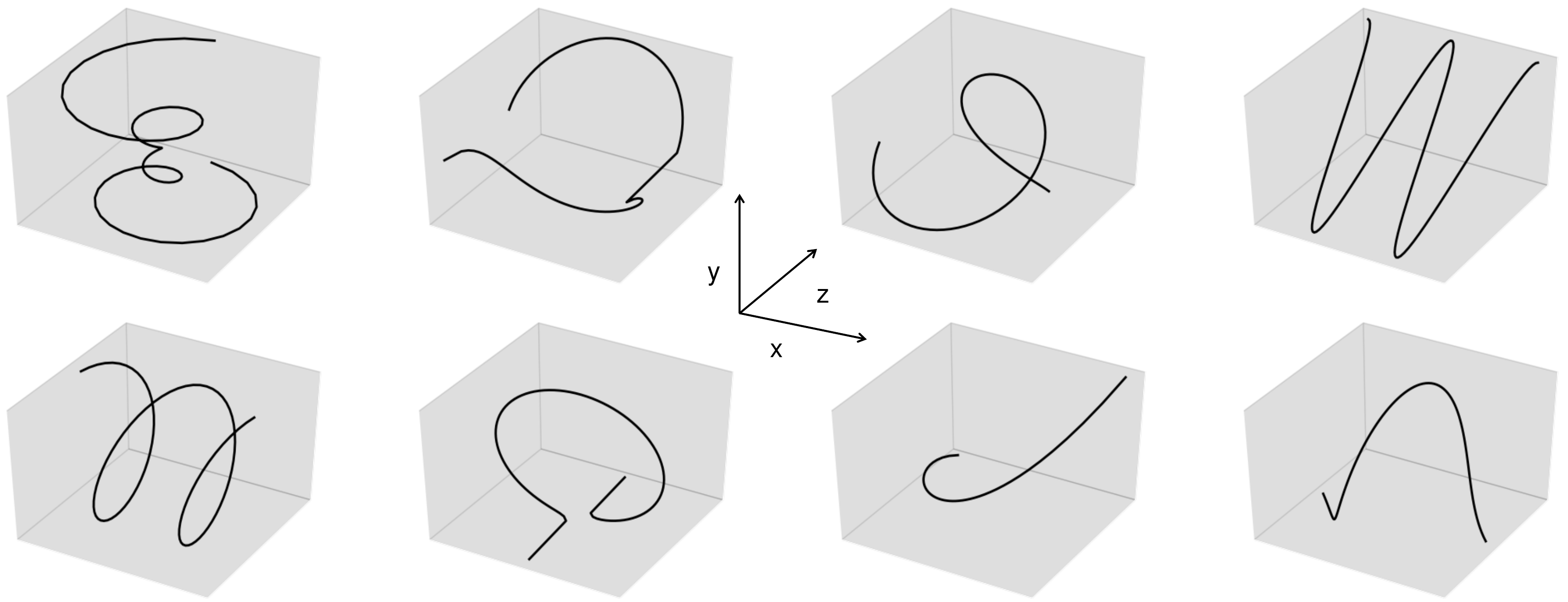}
	\caption{Reference 3D curves.}
	\vspace{-.5cm}
	\label{fig:cruves}
\end{figure}

 The experiments involved a Unity-based 3D trajectory tracking task  \cite{pareek2018position, pareekMyoTrack} wherein the subject uses the end-effector of a robotic/haptic device (Geomagic\textsuperscript{\textregistered} Touch\textsuperscript{TM}) as a writing stylus controller. The subjects were required to track 3D reference trajectories chosen from Fig. \ref{fig:cruves}. Use of 3D trajectories on a 2D computer monitor increased the complexity of the task and led to large (forced) tracking errors despite the use of healthy subjects in the experiment. This was intentional as it forced subjects to make errors and thus enabled the therapist to provide assistance. 

\subsection{Imitation Learning}

 The first experiment was designed to study the imitation learning aspects of iART. During the experiment, as the subjects followed the reference trajectory, the therapist toggled the robotic assistance on or off using a keyboard key press. Assistance was supplied according to the ER rule described in Section \ref{sec:demo}.  Two curves were randomly chosen from Fig. \ref{fig:cruves} to serve as the reference trajectory. These shapes were chosen as they provided a balance between complexity and covering basic motor primitives. The subjects were required to track each trajectory twice. State-action pairs collected during the execution of the first trajectory served as the training dataset $\chi$ for the LSTM network. A different model was trained for each subject. Data from the second trajectory were used for testing the trajectory-agnostic nature of iART. The results from this experiment are summarized in Section \ref{sec:result_imitaiton}. The $K_p$ and $K_d$ gains in (\ref{eq:control}) were set to 4 and 0.001, respectively.

\subsection{Realtime Testing} 

The second experiment tested the realtime prediction capabilities of iART. 7 out of the original 9 subjects were recruited again for this experiment. The trained model obtained from the first experiment was then used for realtime assistance prediction using Python as subjects tracked a reference trajectory chosen from Fig. \ref{fig:cruves}. Local states were streamed to from Unity3D to Python using Transmission Control Protocol. 
Once the assistance was predicted using Python, the assistance behavior was communicated to the robotic device using a simulated key press. During this trial, the therapist used a key press (that had no influence on the robotic assistance) to indicate whenever they would have provided assistance to the user. In other words, while the actual assistance was provided through iART, corresponding assistance behavior of the therapist was recorded simultaneously as  ground truth to verify the realtime performance of iART in mimicking the therapist.



\subsection{DAgger Phase} \label{sec:dagger_testing}

Next, we tested the retraining aspect of iART through the DAgger Phase. The experiment was repeated with one of the subjects. During the demonstration phase, the therapist adopted an 'assist-too-often' strategy wherein, they assisted the subject for very small deviations from the reference trajectory. Additionally, the subject was assisted by the therapist while returning to the starting point (Section \ref{sec:dagger}). An LSTM model was trained using this behavior and a new reference trajectory was chosen for realtime testing and retraining. 

During the DAgger phase, the subject was instructed to modify the assistance behavior of the model using iART's override methodology. They were instructed to retrain the network to prevent it from assisting while the subject navigated to the starting point 
and to assist for larger deviations from the trajectory as opposed to the learned behavior of 'assist-too-often'. 

\subsection{Special Cases}

We designed two special cases to test the imitation learning capabilities of iART using one of the subjects. These case were:
\paragraph{ Assist-too-often} The demonstration phase involved the therapist assisting the subject for small deviations from the reference  trajectory (same as above).
\paragraph{Assist-on-stop} In this case, the therapist assisted the subject whenever they stopped moving the end effector, irrespective of the tracking error.
The models trained using these behaviors were then tested in realtime on different reference trajectories.

\section{RESULTS AND DISCUSSIONS} \label{sec:results}

In this section, we provide the results for the aforementioned experimental evaluation. The efficacy of iART is represented through classification accuracy in the offline and realtime setting and graphical representations to enable visual comparisons between the assistance behavior demonstrated by the therapist and that learnt by iART. We also conducted a paired t-test on the percentage time during which the assistance was switched on by the therapist and iART as a metric to validate the realtime performance of iART. Another statistical analysis was conducted to compare the velocity and position error at which iART and therapist assistance are switched on to understand special behaviors learnt by iART and also to validate the DAgger module.

\subsection{Imitation Learning} \label{sec:result_imitaiton}

Across the nine subjects, the system demonstrated a mean accuracy of $91.41\%$ (range 85.77\%-94.71\%) in predicting (and mimicking) the assistance behavior of the therapist. The system demonstrated a mean accuracy of $90.21\%$ in identifying the assistance-off action (true negatives) and $96.24\%$ in identifying the assistance-on action (true positives).




\begin{figure}[!t]
	\centering
	\includegraphics[ width=.5\textwidth]{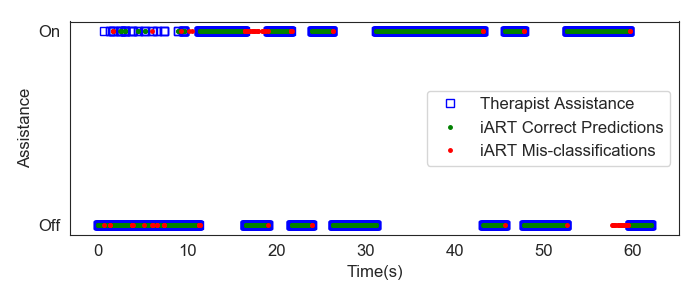}
	\caption{Classification results on one of the test trajectories.}
	\vspace{-.5cm}
	\label{fig:prediction}
\end{figure}

We present the prediction results for one of the test trajectories in Fig \ref{fig:prediction}. Remember that the test trajectory was different from the one used during the demonstration and learning phase. The therapist's actions are denoted by blue squares and the dots represent iART's predicted actions. Cases where iART's actions matched the therapist's are labelled in green and mis-classifications are marked as red. It can be observed that the system performs reliably throughout the experiment. Most mis-classifications occur at the point where the assistance is switched from off to on, and vice-versa. These mis-classifications are usually off by a few time-steps (a few milliseconds) and can be ignored for practical purposes. It is clear from the above observations that iART provides a reliable methodology to mimic a therapist's assistance behavior and also highlights its trajectory agnostic nature.


\subsection{Realtime Testing} \label{sec:result_real}

\begin{figure}[!t]
	\centering
	\includegraphics[width=.5\textwidth]{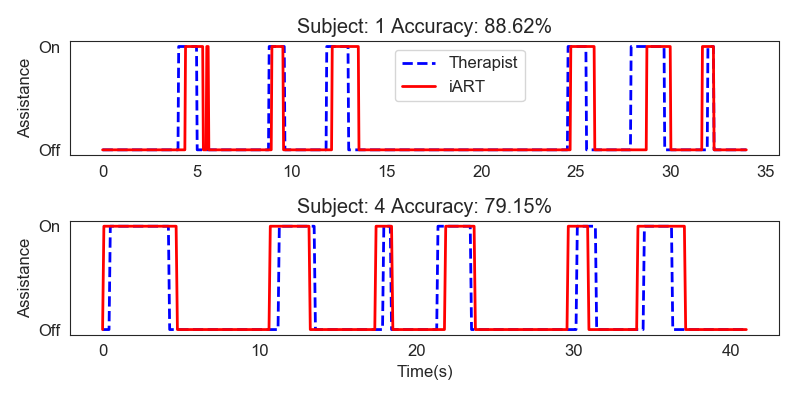}
	\caption{Realtime assistance comparison between therapist and iART for two different subjects.}
	\vspace{-20pt}
	\label{fig:prediction_realtime}
\end{figure}

The system demonstrated a realtime tracking accuracy of 81.85\% (range 76.48\% - 89.31\%) across 7 subjects. The ground truth for this case was collected from the therapist while iART provided the actual assistance. The mean accuracy for these 7 subjects in the offline setting was 88.41\% (Section \ref{sec:result_imitaiton}). Despite the lower accuracy demonstrated by iART in realtime, the assistance pattern was similar to the one exhibited by the ground truth collected from the therapist. 

Fig. \ref{fig:prediction_realtime} demonstrates the performance of iART for two different subjects. In both cases, iART's assistance behavior (solid red line) was similar to the therapist's (dotted blue line). The assistance switching behavior was stable and did not switch from one state to another too often. This enabled a smooth assistance mechanism. 
One difference was the therapist's tendency to switch the assistance on a little earlier when compared with iART. Additionally, iART tends to assist over shorter intervals compared to the assistance provided by the therapist. This may be due to the increased familiarity of the subject with the system and hence the need for shorter bursts of assistance. Since at this point the subject had been using the system for a while, some learning effects may be present, which would have reduced their assistance requirements. 
The subjects also reported that they were unable to distinguish the assistance behavior of iART from the original experiment involving an expert.

We also conducted a paired t-test on the percentage time for which assistance was switched on by the therapist and iART across the 7 subjects. The two demonstrated no significant differences ($t(6) = 0.92$, $p = 0.39$). While the therapist kept the assistance on for an average of 24.81\% (std 5.91\%) of the time across the 7 subjects, iART kept it on for 26.71\% (std 5.44\%). This reaffirms the ability of iART to learn and recreate the therapist's assistance behavior in realtime.

\begin{figure}[b!]
	\centering
	\includegraphics[clip, width=.45\textwidth]{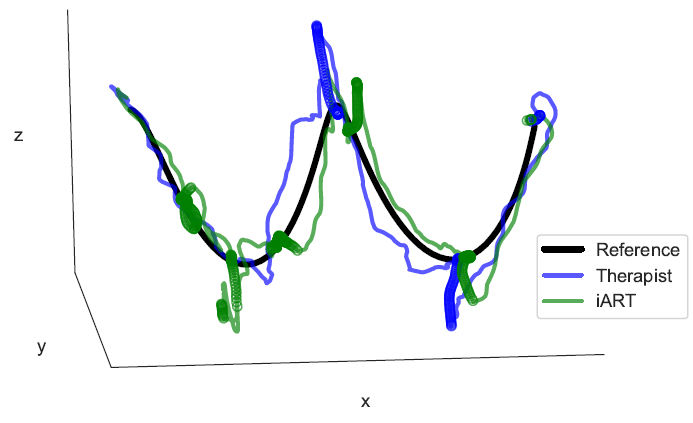}
	\caption{Tracking behavior executed by a subject under therapist and iART assistance for the same reference trajectory.}
	\vspace{-10pt}
	\label{fig:prediction_trajectory}
\end{figure}

We also report iART's performance when tested on the same reference trajectory as used for training. In this case the ground truth refers to therapist assistance collected during the initial demonstration phase. Fig. \ref{fig:prediction_trajectory} demonstrates the tracking behavior exhibited by the subject under therapist (green) and iART (blue) assistance for the same reference trajectory (black). The green and blue circles denote instances where the robotic assistance was switched on by the therapist and iART, respectively. 

iART was able to successfully mimic the therapist's assistance behavior. Additionally, in this particular case, it was observed that the subject required assistance at similar phases of the trajectory. For instance, in both cases the therapist and iART assisted the subject at the mid-way point of the trajectory. Similar behavior was exhibited towards the last quarter of the trajectory.

\begin{figure}[!b]
	\centering
	\includegraphics[trim={0 0 0 0.25cm},clip, width=.48\textwidth]{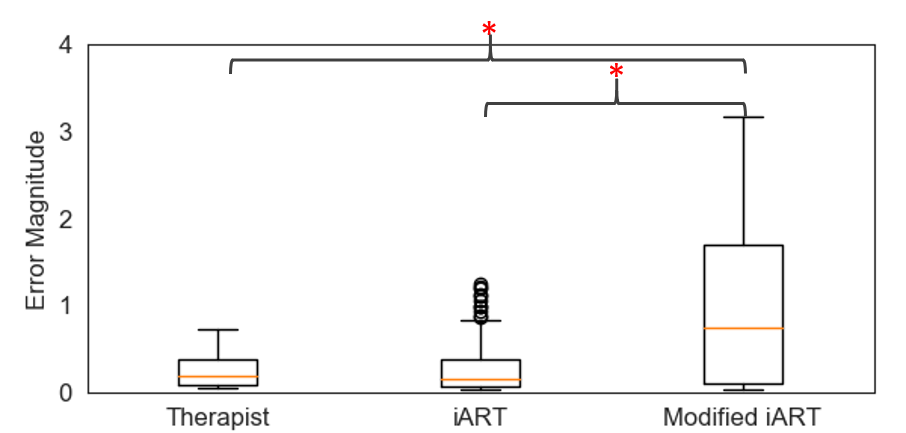}
	\caption{Box plots demonstrating the position error for which the robotic assistance is switched on under: assist-too-often therapist behavior (left); iART's learned assist-too-often behavior (middle); and iART modified using the DAgger phase to assist for larger deviations (right). The red asterisks denote significant differences.}
	\vspace{-.5cm}
	\label{fig:box_plots}
\end{figure}


\subsection{DAgger Phase} \label{sec:results_dagger}

The goal of the experiment was to modify the assist-too-often behavior of the therapist and iART. Fig. \ref{fig:box_plots} shows the box plots for the tracking error magnitude during which the robotic assistance is switched on. The first two plots depict the therapist behavior and the learned iART assistance. It is evident from the plots that both the therapist and iART assist for very small tracking errors as described by the assist-too-often strategy. The final plot describes the modified iART behavior wherein the subject modified the learned iART behavior to assist for larger deviation. 

It was observed that the subject was able to successfully retrain the model to learn the new desired behavior. A Welch's t-test was conducted between the three pairs viz. Therapist-iART, Therapist-Modified iART and iART-Modified iART. Both Therapist-Modified iART and iART-Modified iART demonstrated significant differences at a \textit{p-value} $< 0.02$. reaffirming iART's ability to retrain models. 

Additionally, the subject was instructed to modify the assistance mechanism to not assist as they navigated to the starting point. The subject was able to successfully modify this behavior. The original model assisted for 62.18\% of the time-steps as the subject navigated to the start point. In the modified behavior, this was reduced to 28.25\% of the time-steps.

\subsection{Special Cases} \label{sec:special}

Finally, we report iART's ability to mimic special behaviors exhibited by the therapist. 

\paragraph{Assist-too-often} As explained in the previous section and Fig. \ref{fig:box_plots}, iART can successfully mimic this behavior.

\begin{figure}[!t]
	\centering
	\includegraphics[ width=.5\textwidth]{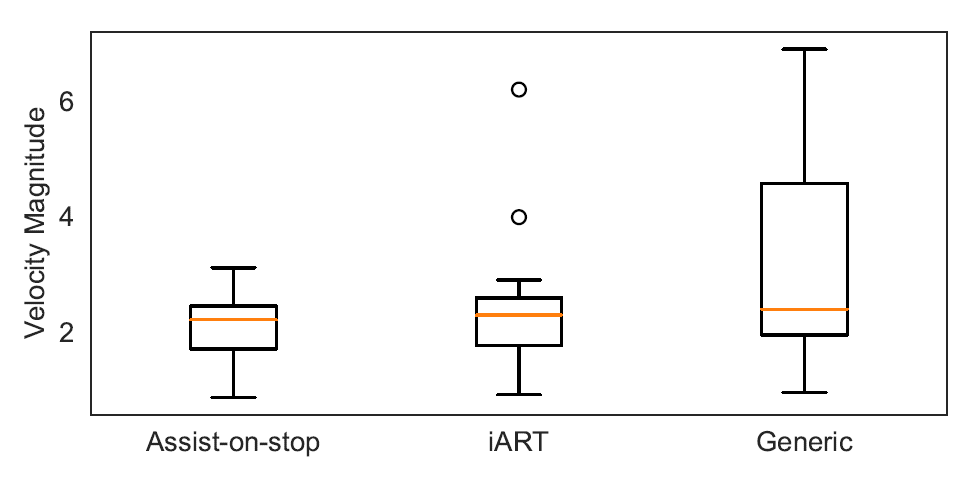}
	\caption{Box plots demonstrating the end-effector velocity at for which the robotic assistance is switched from off to on under different settings.}
 	\vspace{-15pt}
	\label{fig:box_plots_vel}
\end{figure}

\paragraph{Assist-on-stop} For this case, we present the box plots describing the cursor velocity at which the assistance is \textit{toggled from off to on}. Fig. \ref{fig:box_plots_vel} shows this behavior under three different settings. The first plot describes the assistance behavior of the therapist. Remember, the therapist switches on the assistance whenever the end-effector comes to a near stop (low velocity). Compared with the plot for iART, it was observed that both cases exhibited a similar spread of cursor velocity, indicating that iART can successfully learn this behavior. Additionally, we present the assistance start velocity for a generic case wherein assistance was supplied based on the subject's position error and overall performance. Compared to the previous cases, the spread of velocity covers a wider range for this generic case. 



\section{CONCLUSION AND FUTURE DIRECTIONS}

In this paper, we have demonstrated the use of LSTM as a novel paradigm for robot LfD in robotic rehabilitation. We have presented a system capable of mimicking a therapist's assistance behavior with relatively short demonstrations of 2 minutes. As opposed to conventional LfD techniques, our method is trajectory-agnostic and can generalize learned behavior from a single trajectory to any 3D shape. The system can be re-trained based on the preferences of the patient using a modified DAgger-based methodology. iART demonstrates a mean accuracy of $91.41\%$ in predicting, and hence mimicking a therapist's assistance behavior. The system delivers stable performance in realtime and can be retrained easily to modify pre-learnt assistance behaviors.

While the use of local features enables iART to generalize its assistance behavior across multiple trajectories, this generalization fails to capture the critical dynamics that may be associated with a specific shape. Although, the use of radius of curvature aims at capturing this information, it is not the most effective measure and may fail in the presence of obstacles on the path. 

The current system provides assistance in cases wherein subjects deviate from the reference trajectory and is unable to guide subjects requiring assistance along the trajectory. We are in the process of developing a similar LSTM-based methodology that can learn this behavior using therapist demonstrations. The new system would not only predict 'when-to-assist' but will also decide 'how-to-assist' by predicting whether to guide the subject towards the trajectory or along it.


\bibliographystyle{IEEEtran}
\bibliography{IEEEabrv,references}
\appendix \label{appendix}

\begin{figure}[!hb]
	\centering
	\includegraphics[width=.45\textwidth]{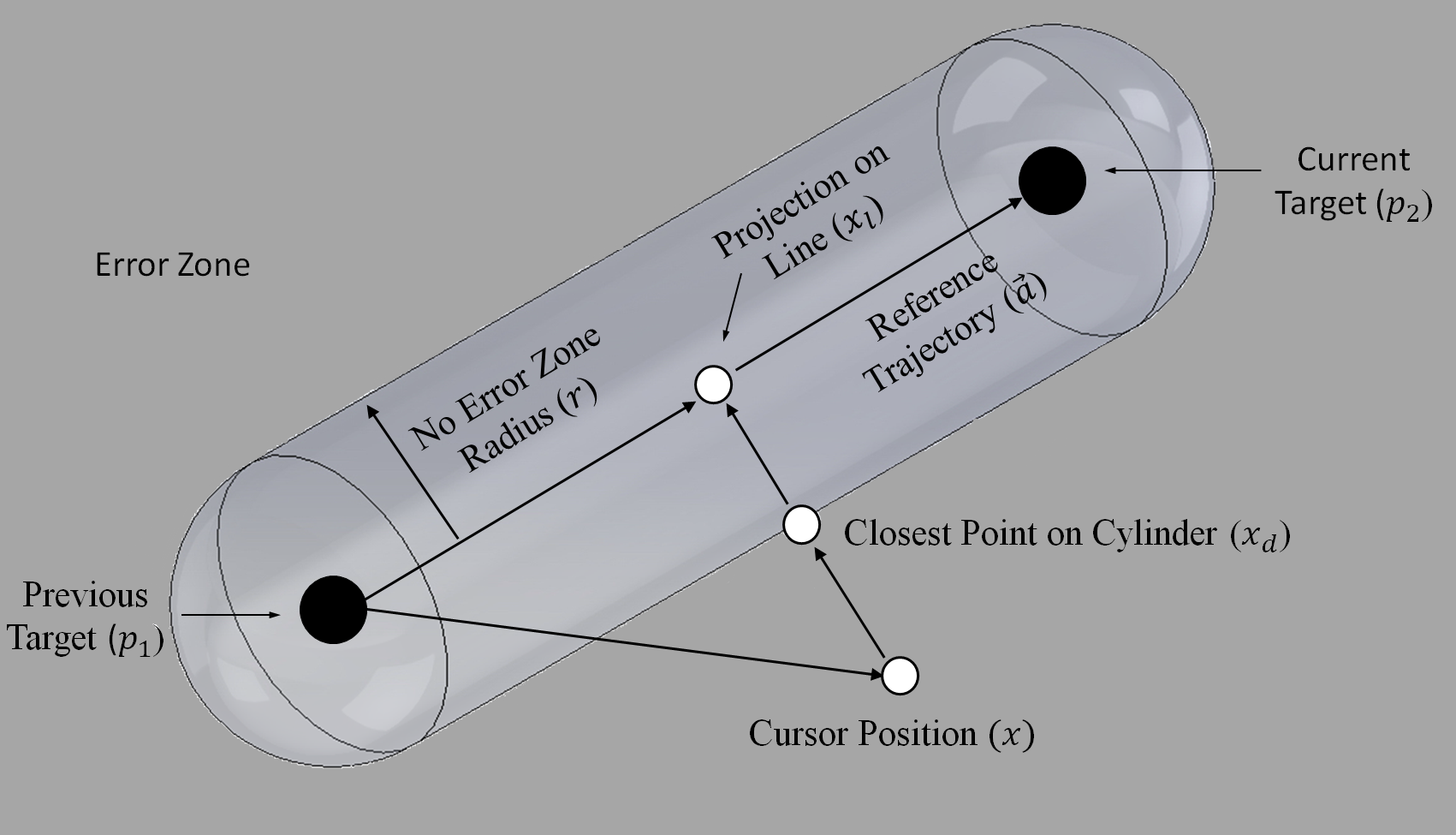}
	\caption{A representation of the virtual no-error zone.} 
	\label{fig:errorZones}
	\vspace{-2mm}
\end{figure}

The error threshold $r$ may be imagined as a virtual 'no-error zone' around the reference trajectory. For instance, the virtual no-error zone for a straight line connecting two points $\mathbf{p_1}$ and $\mathbf{p_2}$ would be a cylinder of radius $r$ (see Fig. \ref{fig:errorZones}). Here, robotic assistance is applied to guide the subject back to the closest point on the cylinder's surface ($\mathbf{x_d}$) using the following rule:


\begin{align}
    \textbf{u}(t)= 
    \begin{cases}
    K_p [\textbf{x}_d(t) - \textbf{x}(t)] - K_d\dot{\textbf{x}}(t), & \text{if } d > r \\
    0,                                          & \text{otherwise}           
    \end{cases}
    \label{eq:control}
\end{align} 

Where, $d$ is the Euclidean distance between the current point $\mathbf{x}(t)$ and closest point on the line $\mathbf{x_l}(t)$. $K_p$ and $K_d$ refer to proportional and derivative gains for the P-d controller.

\addtolength{\textheight}{-12cm} 
\end{document}